\documentclass[fleqn,twoside]{article}

\usepackage[headings]{espcrc2}
\usepackage{algorithm}
\usepackage{algpseudocode}

\readRCS
$Id: espcrc2.tex,v 1.2 2004/02/24 11:22:11 spepping Exp $
\usepackage{graphicx, balance}
\usepackage{epstopdf}
\usepackage{grffile}
\usepackage[figuresright]{rotating}
\usepackage{blindtext}
\usepackage{multicol}
\usepackage{mathrsfs}

\newcommand{\AmS}{{\protect\the\textfont2
  A\kern-.1667em\lower.5ex\hbox{M}\kern-.125emS}}

\usepackage{amsmath}

\title{\textbf{Learning Concept Hierarchies through Probabilistic Topic Modeling}}

\author{V S Anoop\address[DEL]{Data Engineering Lab, Indian Institute of Information Technology and Management - Kerala (IIITM-K), Thiruvananthapuram 695 581,  India, Contact: anoop.res15@iiitmk.ac.in \\},
S Asharaf\address[IIIT]{Indian Institute of Information Technology and Management - Kerala (IIITM-K), Thiruvananthapuram 695 581, India, Contact: asharaf.s@iiitmk.ac.in},
Deepak P\address[QU]{Queen's University, Belfast, UK, Contact:  deepaksp@acm.org}}

\runtitle{Learning Concept Hierarchies through Probabilistic Topic Modeling}
\runauthor{V S Anoop, S Asharaf and Deepak P}

\begin{document}
\begin{abstract}

With the advent of semantic web, various tools and techniques have been introduced for presenting and organizing knowledge. Concept hierarchies are one such technique which gained significant attention due to its usefulness in creating domain ontologies that are considered as an integral part of semantic web. Automated concept hierarchy learning algorithms focus on extracting relevant concepts from unstructured text corpus and connect them together by identifying some potential relations exist between them. In this paper, we propose a novel approach for identifying relevant concepts from plain text and then learns hierarchy of concepts by exploiting subsumption relation between them. To start with, we model topics using a probabilistic topic model and then make use of some lightweight linguistic process to extract semantically rich concepts. Then we connect concepts by identifying an "is-a" relationship between pair of concepts. The proposed method is completely unsupervised and there is no need for a domain specific training corpus for concept extraction and learning. Experiments on large and real-world text corpora such as BBC News dataset and Reuters News corpus shows that the proposed method outperforms some of the existing methods for concept extraction and efficient concept hierarchy learning is possible if the overall task is guided by a probabilistic topic modeling algorithm.  
\end{abstract}

\maketitle

\section{INTRODUCTION}
\label{intro}

Due to rapid growth of text producing and consuming applications, numerous tools and techniques were introduced in the recent past for extracting useful patterns from unstructured text. These patterns are crucial for organizations to discover knowledge out of it and aid in making intelligent decisions. As the amount of such data grows exponentially, already available algorithms performs poor on the scalability and performance aspects. But there are still a lot of avenues where text data is yet to be exploited fully and thus we need new and efficient algorithms to tackle this situation. Platforms such as social networks, e-commerce websites, blogs and research journals generate such data in the form of unstructured text and it is essential to analyze, synthesis and process such data for efficient retrieval of useful information.\\

In text mining, concepts are defined as a sequence of words that constitute real or imaginary entities. Extraction of such entities are non-trivial for applications such as automated ontology generation \cite{ref_1}, document summarization \cite{ref_2} and aspect oriented sentiment analysis \cite{ref_3} to name a few. This is the era of data explosion thus it is very difficult to store, process, manage and most importantly to extract knowledge out of it. To overcome this shortfall, a significant amount of research has been carried out in the recent past for leveraging underlying thematic and semantic structure from text archives. As a result a good number of algorithmic techniques were introduced which are proved to be efficient for the discovery of themes and semantics underlying high dimensional data.\\

Topic Models are suite of text understanding algorithms which statistically generate latent themes pervade a large collection of unstructured text. Since its inception, text mining researchers and practitioners are using it extensively to analyze and organize large document collections. They are unsupervised learning algorithms thus it does not require user tagged corpus to work with. A large number of topic modeling algorithms have been reported in the past with the difference in the assumption they make for modeling topics. Models such as Probabilistic topic models \cite{ref_4} and Latent Dirichlet Allocation (LDA) are some such flavors of topic modeling that attained significant attention.\\

$\textbf{Contributions:}$ This work proposes a novel unsupervised approach for learning concept hierarchies from large unstructured text corpus which is guided by a probabilistic topic modeling approach. To begin with, we model topics from the corpus using Latent Dirichlet Allocation (LDA) algorithm and then uses a lightweight linguistic process to identify concepts which are close to the real world understanding. Then we make use of a subsumption relation \cite{ref_5} ("is-a") to connect concepts which are related thus forms a hierarchy of concepts.\\

$\textbf{Organization:}$ The rest of this paper is organized as follows. We briefly review related works in Section 2. Section 3 introduces the novel approach we have proposed. Detailed explanation of the implementation details is presented in Section 4, and the evaluation of the proposed method is discussed in Section 5. and finally we draw a conclusion and discuss future work in Section 6.

\section{PROBLEM DEFINITION AND RELATED WORK}
\subsection{Problem Definition}
Here, we define the problem formally. Given a large corpus containing unstructured text documents, our problem is to automatically generate concept hierarchies which are close to human understanding. In a nutshell, this paper aims to answer the following research questions :
\begin{enumerate}
	\item Is it possible to automatically extract human interpretable concepts from statistically generated topics using a lightweight linguistic process ?
	\item Can our proposed method learn a hierarchy of such concepts incorporating a subsumption relation between them, which are important in automated ontology generation ?
	\item Given a large but unstructured text corpus, can our topic modeling guided method better extracts and learns concept hierarchies compared to existing algorithms ?
\end{enumerate}
Many recent works have been reported in this direction which proposed many algorithms to extract semantically rich concepts from plain text. In the following section, we due acknowledge some past literatures that discusses methods which are close to our proposed algorithm.

\textbf{Notations used in this paper:} To help narrative, some commonly used notations are shown in Table 1 that are used in the rest of this paper. 
\begin{table}[h]
	\caption{Notations used in subsequent sections}
	\begin{small}
		\begin{tabular}{|r|r|} \hline
			\textbf{Notation} & \textbf{Meaning}  \\ \hline
			tf&term frequency  \\ \hline
			itf &inverse topic frequency   \\ \hline
			$N_{tf}$ & normalized term frequency\\ \hline
			tc & topic collection\\ \hline
			$C_{td}$&topic-document cluster\\ \hline
			MW&muti-word\\ \hline
			$MW_{c}$&multi-word collection\\ \hline
		\end{tabular}
	\end{small}
\end{table}

\subsection{Related Work}
Concept extraction is the process of extracting real or imaginary entities from plain text that has got wider recognition in the recent past. This is due to the wide variety of applications which are mainly dealing with text data such as e-commerce websites, research articles etc. Thus a significant number of research literatures are available in the field of concept extraction and mining which proposes many algorithms with varying degrees of success. In this section, we give emphasis on past literatures in automated concept extraction and hierarchy learning algorithms and briefly discuss works closely related to our proposed framework.\\

Phrase discovering topic model \cite{ref_PDTM} that uses pitman-yor process and TopMine \cite{ref_TopMine} were two notable works that proposed algorithms for mining topical phrases from text documents. The former constructs a topic-word matrix before modeling topics but disadvantage of the approach was that creating such a matrix for large volume of text is often difficult. The latter approach makes use of a two stage process for modeling topics and mainly works with clinical documents. First it identifies phrases using some off-the-shelf tools and then trains a topic model with the identified phrases. Another work which uses topic models for generating multi-word phrases was the topical n-gram \cite{ref_TopicalNGram}. This makes use of some switching variable for identifying a new n-gram. The assumption of this method was that the words within an n-gram usually won't share same topic, which may not be true all the time.\\

Automatic Concept Extractor (ACE), a system specifically designed for extracting concepts from HTML pages and making use of the text body and some visual clues on HTML tags for identifying potential concepts was proposed by Ramirez and Mattmann \cite{ref:ramirez}. Even though this method could outperform some state of the art methods, dependency with HTML was a major drawback. Turney\cite{ref:turney} proposed another system named GenEx, which employed a genetic algorithm supported rule learning mechanism for concept extraction.\\

A system which extracts concepts from user tag and query log dataset is proposed by Parameswaran et.al.\cite{ref:parameswaran} which uses techniques similar to association rule mining. This method uses features such as frequency of occurrences and the popularity among users for extracting core concepts and attempts to build a web of concepts. Even though this algorithm can be applied to any large dataset, a lot of additional processing is required when dealing with web pages.
A bag-of-word approach was proposed by Gelfand et.al.\cite{ref:gelfand} for concept extraction from plain text and used these to form a closely tied semantic relations graph for representing relationships between them. They have applied this technique specifically for some classification tasks and found that their method produces better concepts than the Naive Bayes text classifier. \\

Dheeraj Rajagopal et.al.\cite{ref:dheeraj} introduced another graph based approach for commonsense concept extraction and detection of semantic similarity among those concepts. They used a manually labeled dataset of 200 multi-word concept pairs for evaluating their parser capable of detecting semantic similarity and showed that their method was capable of effectively finding syntactically and semantically related  concepts. The main disadvantage of that method is the use of manually labeled dataset and the creation of such dataset is time consuming and requires human effort. Another work reported in this domain is the method proposed by Krulwich and Burkey \cite{ref:krulwich} which uses a simple heuristics rule based approach to extract key phrases from document by considering visual clues such as the usage of bold and italic characters as features. They have shown that this technique can be extended for automatic document classification experiments. \\

A key phrase extraction system called Automatic Keyphrase Extraction (KEA) developed by Witten et.al\cite{ref:witten} was reported in the concept extraction literatures which creates a Naive Bayes learning model with known key phrases extracted from training documents and uses this model for inferring key phrases from new set of documents. As an extension to this KEA framework, Song et. al.\cite{ref:song} proposed a method which uses the information gain measure for ranking candidate key phrases based on some distance and tf-idf features which was first introduced in \cite{ref:witten}. Another impressive and widely used method was introduced by Frantzi et. al.\cite{ref:frantzi} which extracts multi-word terms from medical documents and named as C/NC method. The algorithm uses a POS tagger POS patten filter for collecting noun phrases and then uses some statistical measures for determining the termhood of candidate multi-words. \\

The proposed method in this paper is a hybrid approach incorporating statistical methods such as topic modeling and tf-itf weighting and some lightweight linguistic processes such as POS tagging and analysis for leveraging concepts from text. We expect the learnt concept hierarchy to be close to the real world understanding of concepts which we will quantify using evaluation measures such as precision, recall and f-measure.

\section{BACKGROUND : LATENT DIRICHLET ALLOCATION (LDA)}
A good number of topic modeling algorithms are introduced in the recent past which varies in their method of working mainly with the assumptions they adopt for the statistical processing. An automated document indexing method based on a latent class model for factor analysis of count data in the latent semantic space has been introduced by Thomas Hofman \cite{ref:hofman}. This generative data model called Probabilistic Latent Semantic Indexing (PLSI), considered as an alternative to the basic Latent Semantic Indexing has a strong statistical foundation. The basic assumption of PLSI is that each word in a document corresponds to only one topic.\\

Later, Blei et. al.\cite{ref:blei} introduced a new topic modeling algorithm known as Latent Dirichlet Allocation (LDA) which is more efficient and attractive than PLSI. This model assumes that a document contain multiple topics and such topics are leveraged using a Dirichlet Prior process. In the following section, we will briefly describe the underlying principle of LDA. Even though a LDA works well on broad ranges of discrete datasets, the text is considered to be a typical example to which the model can be best applied. The process of generating a document with $n$ words by LDA can be described as follows\cite{ref:blei}:
\begin{enumerate}
	\item Choose the number of words, $n$, according to Poisson Distribution;
	\item Choose the distribution over topics, $\theta$, for this document by Dirichlet Distribution;
	\begin{enumerate}
		\item Choose a topic $T^{(i)}$ $\sim$ Multinomial$(\theta)$  
		\item Choose a word $W^{(i)}$ from $P\left ( W^{(i)} |  T^{(i)} , \beta \right )$
		\end{enumerate}
		\end{enumerate}
		Thus the marginal distribution of the document can be obtained from the above process as :\\
		\begin{equation*}
		\begin{aligned}
		P\left ( d \right ) =  \int_{\theta} \left ( \prod_{i=1}^{n}\sum_{T^{(i)}} P(W^{(i)}|T^{(i)}, \beta ) P(T^{(i)}|\theta)\right )\\
		+ P(\theta|\alpha)d\theta
		\end{aligned}
		\end{equation*}
		 
where, $P(\theta | \alpha)$ is derived by Dirichlet Distribution parameterized by $\alpha$, and $P(W^{(i)})|T^{(i)}, \beta )$ is the probability of $W^{(i)}$ under topic $T^{(i)}$ parameterized by $\beta$. The parameter $\alpha$ can be viewed as a prior observation counting on the number of times each topic is sampled in a document, before we actually seen any word from that document. The parameter $\beta$ is a hyper-parameter determining the number of times words are sampled from a topic \cite{ref:blei}, before any word of the corpus is observed. At the end, the probability of the whole corpus $D$ can be derived by taking the product of all documents' marginal probability as given below:
		\begin{equation}\label{eq:third}
		P(D) = \prod_{i = 1}^{M} P(d_i)
		\end{equation}
		
\section{PROPOSED APPROACH}
In the area of text mining, topic models or specifically probabilistic topic models are suite of algorithms which got wider recognition for its ability to leverage hidden thematic information from huge archives of text data. Text mining researchers are making use of topic modeling algorithms such as Latent Semantic Analysis (LSA) \cite{ref_LSA}, Probabilistic Latent Semantic Indexing (pLSI) \cite{ref_PLSI}, Latent Dirichlet Allocation (LDA) \cite{ref_LDA} etc extensively for bringing out the themes or so called "topics" from high dimensional unstructured data. \\

Among all these algorithms, LDA has got lot of attention in the recent past and is widely using because of its easiness of implementation and potential applications. Even though the power of LDA algorithm has been extensively used for leveraging topics, very few studies have been reported for mapping these statistically outputted topics to semantically rich concepts. Our proposed framework is an attempt to address this issue by making use of LDA algorithm to generate topics and we leverage concepts from such topics by using a new statistical weighting scheme and some lightweight linguistic processes. The overall work flow of the proposed approach is depicted in Fig.1.\\

\begin{figure*}[t]
	\centerline{\includegraphics[width=350px,height=125px]{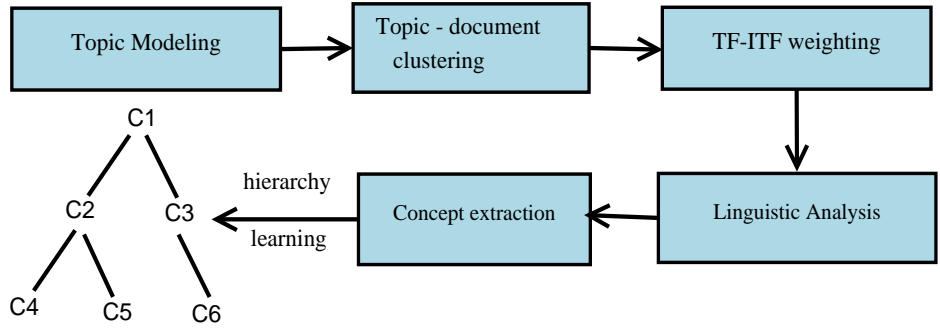}}
	\caption{Workflow of the proposed framework, where C1, C2, C3, C4, C5 and C6 represent concepts\vspace*{-6pt}}
\end{figure*}

Our framework can be divided into 2 modules (i) concept extraction and (ii) concept hierarchy learning. The concept extraction module extract concepts from topics generated by LDA algorithm and the concept hierarchy learning module learns a hierarchy of extracted concepts by inducing a subsumption hierarchy learning algorithm. Detailed explanation of these modules are given below.

\subsection{Concept Extraction}
In this module, we introduce a topic to concept mapping procedure for leveraging potential concepts from statistically computed topics which are generated by the LDA algorithm. The first step of the proposed framework deals with the preprocessing of data which is meant for removing unwanted and irrelevant data and noises. Latent Dirichlet Allocation algorithm is executed on top of this preprocessed data which in turn generate topics through the statistical process. A total of 50 topics have been extracted by tuning the parameters of LDA algorithm. Once we got the sufficient topics for the experiment, for each topic, we have created a topic - document cluster by grouping the documents which generated such a topic and the same process has been executed for all topics under consideration.\\

Now, we introduce a new weighting scheme called $tf-itf$ (term frequency - inverse topic frequency) which is used for finding out highly contributing topic word in each topic. We bring this weighting scheme to filter out the relevant candidate topic words. Term frequency ($tf$) is the total number of times that particular topic word comes in the topic - document clusters. Normalized term frequency, $N_{tf}$ of a topic word $T_w$ can be calculated as:
\begin{equation}
N_{tf} = \frac{count(T_w)~in~C_{td}}{count(total~terms~in~C_{td})}
\end{equation}
Inverse topic frequency $I_{tf}$ is calculated as:
\begin{equation}
I_{tf} = \frac{count(total~terms~in~C_{td})}{count(documents~with~T_w)}
\end{equation}
$tf-itf$ is calculated using the following equation:
\begin{equation}
tf-itf = 	N_{tf} * I_{tf}   
\end{equation}
This step is followed by a sentence extraction process in which all the sentences which contain the topic words which have high tf-itf weight are extracted. Next, we apply a parts of speech tagging on these sentences and extract only noun and adjective tags as we are only concentrating on the extraction of concepts. In linguistic pre-processing step, we take Noun + Noun, Noun + Adjective and (Adjective / Noun) + Noun combinations of words from the tagged collection. Concept identification is the last step in the process flow in which we find out the term count of all the combinations of Noun + Noun, Noun + Adjective and (Adjective / Noun) + Noun. A positive term count implies that the current multi word can be a potential "concept" and if we get a zero term count, then that multi word can be ignored. The newly proposed algorithm for extracting the concepts is shown in Algorithm 1. 

\begin{algorithm}
	\caption{Concept Extraction}
	\label{conceptextraction}
	\begin{algorithmic}[1]
		\Procedure{ExtractConcepts}{$tc$}
		\State $\forall$ $t$, create $C_{td}$
		\State $\forall C_{td}$, compute $tf-itf$ weight
		\State $\forall$ $t$, choose $n$ words with highest $tf-itf$
		\State $S[~]$ = sentences with top $tf-itf$ words
		\State $POS~tag(S)$  
		\State $W[~]$ = $(NNP, NNS, NN, JJ)$
		\State $MW_c[~]$ = $noun+noun | adj+noun$
		\While{$|MW_c|\not=0$}
		\State $termCount(MW)$ $\forall$ $MW$ in $MW_c$
		\If{$T_c > 0$}
		\State Add $MW$ into $C$
		\State Remove $MW$ from $MW_c$
		\State Fetch next $MW$ from $MW_c$
		\Else
		\State Remove $MW$ from $MW_c$
		\State Fetch next $MW$ from $MW_c$
		\EndIf
		\EndWhile
		\EndProcedure
	\end{algorithmic}
\end{algorithm}

\subsection{Concept Hierarchy Learning}
In this module we derive hierarchical organization of leveraged concepts using a type of co-occurrence called "subsumption" relation. Subsumption relation is found to be simple but very effective way of inferring relationships between words and phrases without using any training data or clustering methods. The basic idea behind subsumption relation is very simple : for any two concepts $C_a$ and $C_b$, $C_a$ is said to be subsume $C_b$ if 2 conditions hold.
$P(C_a|C_b) = 1$ and $P(C_b|C_a) < 1$. To be more specific, $C_a$ subsumes $C_b$ if the documents which $C_b$ occurs in are a subset of the documents which $C_a$ occurs in. Because $C_a$ subsumes $C_b$ and because it is more frequent, in the hierarchy, $C_a$ is the parent of $C_b$.
\begin{algorithm}[h]
	\caption{Concept Hierarchy Learning}
	\label{learning}
	\begin{algorithmic}[1]
		\Procedure{LearnHierarchy}{$C$}
		\State Choose pair of concepts, say $C_a$ and $C_b$
		\State Compute $P(C_a|C_b)$ and $P(C_b|C_a)$
		\If{$P(C_a|C_b) = 1$ and $P(C_b|C_a) < 1$}
		\State Assign $C_a$ as the parent of $C_b$  
		\Else
		\State Fetch next concept pairs
		\EndIf
		\State Goto step 2, repeat $\forall$ $C_a$, $C_b$
		\EndProcedure
	\end{algorithmic}
\end{algorithm}

\section{EXPERIMENTAL SETUP}
This section concentrates on the implementation details of our proposed framework and concept extraction and hierarchy learning procedures are discussed in detail.

\subsection{Concept Extraction}
Here, concept extraction module of the framework is discussed. This module concentrates on tasks such as data collection and pre-processing, topic modeling, topic-document clustering, tf-itf weighting, sentence extraction and POS tagging, linguistic pre-processing etc for identifying concepts and a detailed explanation of each step is given below.

\subsubsection{Dataset Collection and Pre-processing}
We are using publicly available datasets such as Reuters Corpus Volume 1 dataset\cite{ref:reuters} and BBC News Dataset\cite{ref:bbc} for the experiment. Reuters is the world's biggest international news agency and cater different news and related information through their website, video, interactive television and mobile platforms. Reuters Corpus Volume 1 is in XML format and is freely available for research purpose. Text messages are extracted by a thorough  pre-processing such as removing XML tags, URLs and other special symbols and then created a new dataset exclusively for our experiment. BBC provides two benchmarked news article datasets which is freely available for machine learning research. The general BBC dataset consist of 2225 text documents directly from their website corresponding to stories in five areas such as business, entertainment, politics, sports and technology, from 2004 to 2005. A thorough pre-processing such as stemming, and removal of stop-word, URLs and special characters on this dataset and made an experiment ready copy of the original dataset.

\subsubsection{Topic Modeling}
Latent Dirichlet Allocation (LDA) algorithm has been applied on the pre-processed dataset to leverage topics for this experiment. The number of iterations is set to 300 as Gibbs sampling method usually approaches the target distribution after 300 iterations. The number of topics is set to 50 and a snapshot of 5 topics we have randomly chosen is shown in Table 2. 

\begin{table}[h]
	\renewcommand{\baselinestretch}{1}
	\caption{Top 10 topic words from first 4 topics along with their TF-ITF weight}
	\begin{center}
		\begin{tabular}{|r|r|} \hline
			\textbf{Topic 1}  & \textbf{Topic 3} \\ \hline
			web [0.0048]  		& set [0.0047] 		  \\ \hline
			search [0.0048]  	& software [0.0032] \\ \hline
			online [0.0047]  	& virus [0.0028]  	  \\ \hline
			news [0.0046]  		& users [0.0027]    \\ \hline
			google [0.0033]  	& firms [0.0025]    \\ \hline
			people [0.0032]  	& microsoft [0.0025]   \\ \hline
			information [0.0032]& security [0.0022]    \\ \hline
			internet [0.0029]  	& windows [0.0022]    \\ \hline
			website [0.0027]  	& file [0.0013]    \\ \hline
			users [0.0020] 		& programs [0.0011]  \\ \hline
			\textbf{Topic 2}  & \textbf{Topic 4} \\ \hline
			system [0.0064]  	& site [0.0042]   \\ \hline
			music [0.0045]  	& net [0.0038]  	\\ \hline
			devices [0.0043]  	& spam [0.0035]  		\\ \hline
			players [0.0035]  	& mail [0.0028]  	\\ \hline
			media [0.0032]  	& firm [0.0027]			\\ \hline
			digital [0.0027] 	& data [0.0024]  			\\ \hline
			market [0.0024] 	& attacks [0.0019]  	\\ \hline
			technology [0.0022]	& network [0.0018]  	\\ \hline
			consumer [0.0021]  	& web [0.0016]  		\\ \hline
			technologies [0.0018] & research [0.0014] 		\\ \hline
		\end{tabular}
	\end{center}
\end{table}

\subsubsection{Topic - Document Clustering}
In this step, we consider each topic and then grouped and clustered top 50 documents which contributed the creation of that specific topic. This has been done for all the 50 topics of our choice. As an outcome, we have got 50 such clusters that contain documents which generated the topics.

\subsubsection{TF-ITF Weighting}
Here, we compute the $tf-itf (term~frequency~-~inverse~topic~frequency)$ weight of each word in every topic using Eq.(3), Eq.(4) and Eq.(5) to find out highly used topic words in the collection. Table 2 also shows topic words along with their tf-itf weight.
\begin{table}[h]
	\caption{Top concepts extracted against first 4 topics}
	\begin{small}
	\begin{tabular}{|r|r|} \hline
		\textbf{Concepts\_Topic 1} & \textbf{Concepts\_Topic 3}\\ \hline
		web search & music players\\ \hline
		search engine & digital media \\ \hline
		google news & digital technology \\ \hline
		online news search & consumer devices \\ \hline
		google search engine & market system\\ \hline
		\textbf{Concepts\_Topic 2} & \textbf{Concepts\_Topic 4} \\ \hline
		software users & spam mail\\ \hline
		virus programs & spam website\\ \hline
		windows security & network research\\ \hline
		software forms & research firm \\ \hline
		microsoft programs & website attacks \\ \hline
	\end{tabular}
	\end{small}
\end{table}

\subsubsection{Sentence Extraction \& POS Tagging}
In sentence extraction step, we consider topic words having highest tf-itf weight and then extract sentences containing these topic words from the topic - document clusters. Then a parts of speech tagging has been done to identify words tagged as nouns and adjectives from these sentences as our aim is to extract potential "concepts" from the repository. For this experiment, Natural Language Toolkit (NLTK) \cite{ref:nltk} has been used which contains libraries for Natural Language Processing for Python programming language.  

\subsubsection{Linguistic Processing \& Concept Identification}
All words which are tagged as Nouns(NN/NNP/NNS) and Adjectives (JJ) are filtered out and all possible combinations of $Noun + Noun, Adjective + Noun$ and $(Noun / Adjective) + Noun$. The results are shown in Table 3. The term count for each of these multi word term is then calculated against the original corpus and a positive term count implies that the corresponding multi-word term can be a potential concept and we eliminate the term if we get a zero term count. This process has been repeated for all the multi-words we have filtered out.

\subsection{Concept Hierarchy Learning}
Concept hierarchy learning module concentrates on leveraging a subsumption hierarchy\cite{ref_5} depicting an "is-a" relation between the concepts identified by the proposed algorithm. Subsumption relation is simple but considered as an important relationship type in any ontological structure and we calculate two probability conditions for the same. For any given two concepts, we first calculate $P(C_1 | C_2)$ and then $P(C_2 | C1)$, in order to establish a subsumption relation, the former probability must be 1 and the latter should be less than 1. In other words, $C_1$ subsumes $C_2$ if the documents in which $C_2$ occurs is a subset of the documents which $C_1$ occurs in. \\

For instance, consider two concepts \textit{dial-up internet} and \textit{network connection}, the proposed method computes
$P(dialup~internet|network~connection)$  and $P(network~connection|dial-up~internet)$ and found that the number of documents in which $dialup~internet$ occurs is a subset of number of documents in which $network~connection$ occurs. That means there exists a subsumption relation between these two concepts and   $dialup~internet$ concept may be subsumed by $network~connection$ concept. This process has been repeated for all concepts in the collection, and a part of such a hierarchy generated using our proposed algorithm is shown in Fig. 2.
\begin{figure}[t]
	\centerline{\includegraphics[width=200px,height=200px]{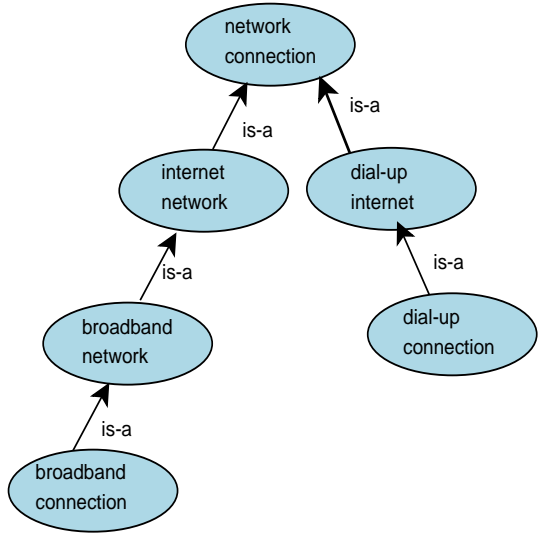}}
	\caption{Part of a subsumption hierarchy learned using Algorithm 2}
\end{figure}

\section{EVALUATION OF RESULTS}
Here we evaluate the results produced by our proposed method and precision and recall measures are used for evaluating the quality of concepts leveraged. We have first created a human generated concept repository and kept for verifying against the machine generated concepts. Precision computes the fraction of machine extracted concepts that are also human generated, and recall measures concepts which are extracted by proposed algorithm that are also human authored. In information retrieval, it is estimated that achieving high precision and recall at same time is difficult and using a measure called F1, we can balance these two. Here, true positive is defined as the number of overlapped concepts between human generated concepts and concepts extracted by our proposed algorithm, false positive is the number of extracted concepts that are not truly human authored concepts and false negative is the human authored concepts that are missed by the concept extraction method. Using these measures, we have compared our proposed method against some of the existing concept extraction algorithms and the result is shown in Table 4. \\

From the performance graph shown in Figure 4, it is clear that our proposed algorithm extracts more concepts as the number of topics are increasing. The other baseline algorithms such as ACE and ICE performs poor when the number of topics are increased randomly. This shows that the proposed algorithm outperforms the baseline algorithms when extracting real-world concepts from large number of statistically generated topics.\\

\begin{table}[h]
	\caption{Comparison of ACE, ICE and our proposed method}
	\begin{small}
	\begin{tabular}{|r|r|r|r|} \hline
		Algorithm & Precision & Recall & F1\\ \hline
		ACE &0.2372 &0.2689 &0.2517 \\ \hline
		ICE &0.7113 &0.8147 &0.7595 \\ \hline
		\textbf{Proposed} &\textbf{0.8165} &\textbf{0.8901} &\textbf{0.8516}  \\ \hline
	\end{tabular}
\end{small}
\end{table}

\begin{figure}[t]
	\centerline{\includegraphics[width=200px,height=150px]{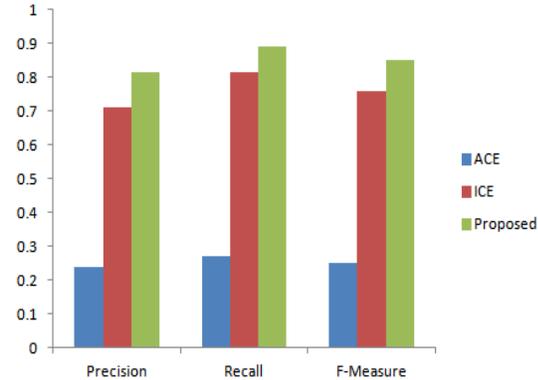}}
	\caption{Precision, Recall and F-measure comparison of ACE, ICE and proposed method}
\end{figure}

\begin{figure}[t]
	\centerline{\includegraphics[width=200px,height=150px]{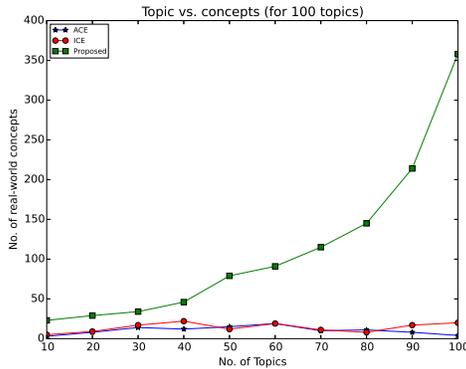}}
	\caption{Performance of algorithms on 100 topics (No. of human interpretable concepts generated)}
\end{figure}

\section{CONCLUSIONS AND FUTURE WORK}
This paper proposed a novel framework for extracting close to real world concepts from large collection of unstructured text documents which is guided by a probabilistic topic modeling algorithm. Proposed method also deals with learning a subsumption hierarchy which exploits "is-a" relationships among identified concepts which is extensively used in ontology generation. Experiments conducted on large datasets such as Reuters and BBC news corpus shows that the proposed method outperforms some of the already available algorithms and better concept identification is possible with this framework.\\

Because of the promising end results, we are interested to work mainly on the directions of measuring the scalability of proposed framework by using more large datasets. Apart from the basic subsumption hierarchy which depicts "is-a" relation, our future work will be on leveraging other relations that exist between concepts we would like to  so that a this framework can automate the complete ontology generation process.

\newpage
\noindent{\includegraphics[width=1in,height=1.7in,clip,keepaspectratio]{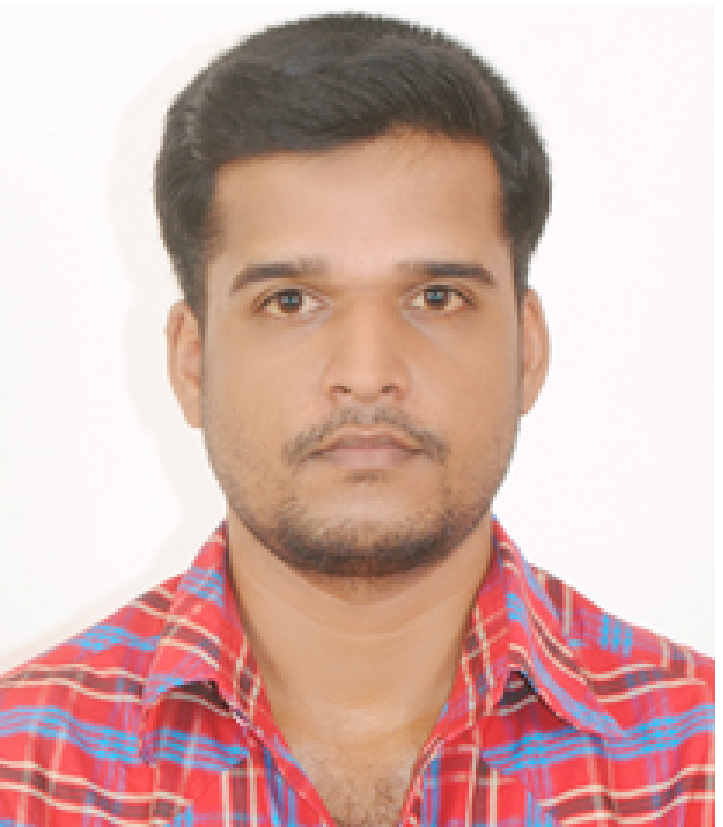}}
\begin{minipage}[b][1in][c]{1.8in}
	{\centering{\bf{Anoop V S }}  is a full time Ph.D Research Scholar at Data Engineering Lab, Indian Institute of Information Technology and Management - Kerala (IIITM-K), Thiruvananthapuram, India. He received his Masters in Computer Applications (MCA) - }\\\\\\
\end{minipage}
from IGNOU and Master of Philosophy  in Computer Science from Cochin University of Science and Technology (CUSAT), Kerala in 2014. He has several publications in international journals, conference proceedings and book chapters. His research interests include Information Retrieval, Text Mining and NLP.\\\\\\\\

\noindent{\includegraphics[width=1in,height=1.7in,clip,keepaspectratio]{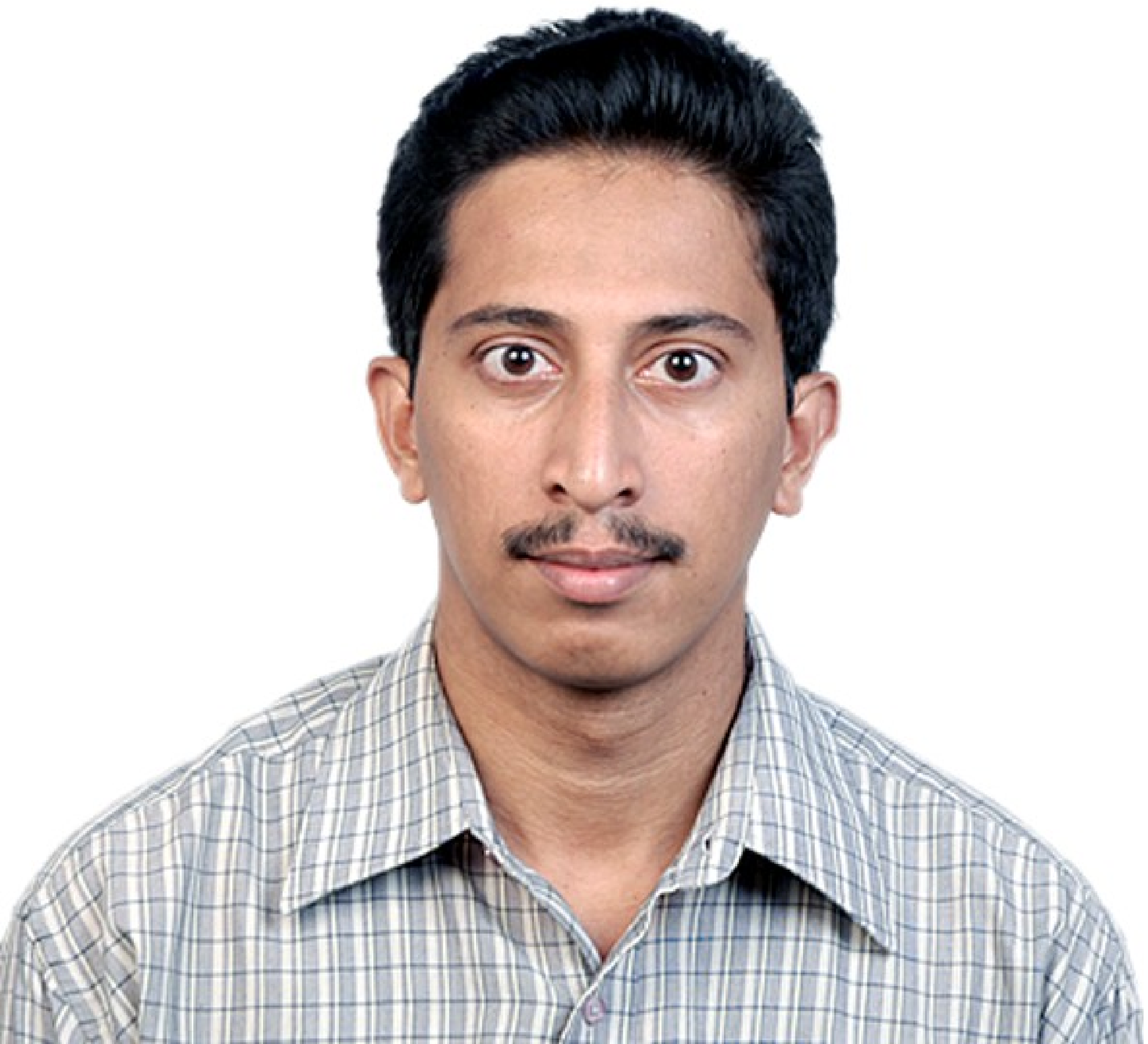}}
\begin{minipage}[b][1in][c]{1.8in}
	{\centering{\bf{Asharaf S }}  is an Associate Professor at Indian Institute of Information Technology and Management - Kerala (IIITM-K), Thiruvananthapuram, India. He received his Ph.D and Master of Engineering degrees in Computer Science and Engineering -  }\\\\\\
\end{minipage}
   from Indian Institute of Science, Bangalore. His areas of interest include algorithms, business models and software systems related to data mining, data analytics, information retrieval, computational advertising, soft computing and machine learning.\\\\

\noindent{\includegraphics[width=1in,height=1.7in,clip,keepaspectratio]{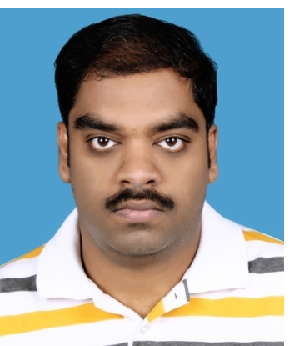}}
\begin{minipage}[b][1in][c]{1.8in}
	{\centering{\bf{Deepak Padmanabhan}} is a Lecturer (Asst. Professor) in Computer Science at Queen’s University Belfast, UK. He completed his M.Tech and PhD from Indian Institute of }
\end{minipage}
  Technology Madras, all in Computer Science. His current research interests include data analytics, similarity search, information retrieval and natural language processing. He has published over 40 research papers across major venues in Information and Knowledge Management. He is a Senior Member of the IEEE and ACM.
\end{document}